\documentclass[10pt,twocolumn,letterpaper]{article}

\usepackage{iccv}
\usepackage{times}
\usepackage{epsfig}
\usepackage{graphicx}
\usepackage{amsmath}
\usepackage{amssymb}
\usepackage{multirow}
\usepackage[normalem]{ulem}

\usepackage[pagebackref=false,breaklinks=true,letterpaper=true,colorlinks,bookmarks=false]{hyperref}

\iccvfinalcopy %

\def\iccvPaperID{7148} %

\ificcvfinal\fi

\usepackage[dvipsnames]{xcolor}
\definecolor{first}{RGB}{248,158,156}
\definecolor{second}{RGB}{247,206,160}
\definecolor{third}{RGB}{255,255,166}

\usepackage{colortbl}

\begin{document}

\title{Graph-based Asynchronous Event Processing for Rapid Object Recognition}

\author{Yijin Li, Han Zhou, Bangbang Yang, Ye Zhang, Zhaopeng Cui, Hujun Bao, Guofeng Zhang\thanks{Corresponding author: Guofeng Zhang.} \\ 
State Key Lab of CAD\&CG, Zhejiang University\thanks{The authors except Zhaopeng Cui are also affiliated with ZJU-SenseTime Joint Lab of 3D Vision. This work was partially supported by NSF of China (No. 61822310 and 61932003).}
}

\maketitle
\ificcvfinal\thispagestyle{empty}\fi

\begin{abstract}
    Different from traditional video cameras, event cameras capture asynchronous events stream in which each event encodes pixel location, trigger time, and the polarity of the brightness changes. In this paper, we introduce a novel 
    graph-based framework for event cameras, namely SlideGCN.
    Unlike some recent graph-based methods that use groups of events as input, our approach can efficiently process data event-by-event, unlock the low latency nature of events data while still maintaining the graph's structure internally. For fast graph construction, we develop a radius search algorithm, which better exploits the partial regular structure of event cloud against k-d tree based generic methods. 
    Experiments show that our method reduces the computational complexity up to 100 times with respect to current graph-based methods while keeping state-of-the-art performance on object recognition.
    Moreover, we verify the superiority of event-wise processing with our method. When the state becomes stable, we can give a prediction with high confidence,  thus making an early recognition. Project page: \url{https://zju3dv.github.io/slide_gcn/}.
\end{abstract}

 \section{Introduction}

Rapid object recognition is essential for a variety of applications, such as autonomous driving and flying drones. For instance, when an autonomous vehicle is driving at high speed, the low latency is desirable to identify obstacles or moving %
objects once they appear. 
Due to its low frame rate, the standard video camera is not ideal for this task. Fast-speed video cameras can have more than 1000 frames per second, while they are normally very expensive and the information is also highly redundant.
As a result, event cameras~\cite{berner2013240, posch2010qvga, indiveri2015neuromorphic} attracts more attention recently due to their high temporal resolution and low latency (both in the order of microseconds) as well as high dynamic range without motion blur.  Compared with video cameras that output images with a specific frame rate, event cameras are event-driven. When a certain brightness change occurs on a pixel, the event camera will trigger an individual event. 
In this way, they naturally discard redundant information by only measuring brightness.

However, since the output of an event camera is a sparse asynchronous events stream, existing efficient methods~\cite{howard2017mobilenets, tan2019efficientnet} which typically work on frames can not be directly applied for event cameras.
As a result, most works~\cite{GehrigLDS19est, CanniciCRM20matrix-lstm, Rebecq19cvpre2vid} transform such events stream to regular 2D event frames or 3D voxel grids before processing.
However, these data representation transformations discard the sparsity of events data and quantify event timestamps, which are likely to obscure the natural invariance of the data
Another type of approach is directly tailored to the sparse and asynchronous nature of event-based data.
Time-surface-based methods~\cite{LagorceOGSB17HOTS, SironiBBLB18HATS} and Spiking Neural Networks (SNNs)~\cite{OrchardMEPTB15sipiking_hfirst, LiuRXT020spiking_recognition, AmirTBMMNNAGMKD17spiking_gesture} are two dominant classes of methods for event-by-event processing. Despite keeping low latency, both methods have limited accuracy in high-level tasks, mainly due to their sensitivity to tuning and difficulty in the training procedure, respectively.
To fully utilize the spatial-temporal sparsity of event data, some recent methods~\cite{wang2019gesture, sekikawa2019eventnet,bi2019graph,mitrokhin2020learning} introduce a compact graph representation that interprets an event sequence as a graph on event cloud and employs graph convolutional networks.
Although these graph-based methods, \eg, \cite{bi2019graph, mitrokhin2020learning}, reach state-of-the-art performance, they rely on integrating events over a certain number of events or events within a period.
They gather the information contained in groups of events at the cost of discarding the low latency nature of events data.

Based on all these observations, in this paper, we propose a novel graph-based recursive algorithm with a sliding window strategy that can process the stream event-by-event efficiently while maintaining high accuracy.
However, it is non-trivial to apply the sliding window strategy for graph-based and event-wise processing.
The naive sliding window strategy is inefficient because it needs to process all the nodes in the graph even with a minor change, although many nodes' features don't change. Moreover, graph construction is prerequisite for graph neural networks, and the radius search is normally adopted \cite{bi2019graph, mitrokhin2020learning} to determine nodes' connection, which is very slow. 
Take the k-d tree based search as an example, 
frequent insertion and deletion will make it unbalanced and cause query performance to drop while rebuilding the index will bring the extra cost to insertion.

To solve these problems, we first propose a novel incremental graph convolution, namely slide convolution, that exploits the local spatial connectivity of convolution and reuses previous calculations in order to avoid processing all nodes. For a single layer, it is rather simple to just compute the features around %
newly added nodes. For multiple-layer GCN, we need to solve the propagation of modified features between layers with different graph topologies. Thus we derive a series of propagation rules.
In this way, we reduce the computational complexity up to 100 times in comparison with the naive sliding window strategy.
Moreover, considering that events locate in the image grid (which consists of two limited and discrete dimensions) rather than generic 3D continuous metric spaces, we introduce a novel radius search algorithm 
for the structure of event cloud, cutting the search cost by half and reducing the cost of insertion and deletion operations to O~(1).

A straightforward application of event-wise processing is early object recognition, as when enough information is received, the prediction result becomes stable and it is not necessary %
to process more events.
Previous works either focus on how to process event-by-event efficiently or reach a certain level of accuracy with less information, but lack the ability of early recognition.
In this paper, we further apply our graph-based recursive method to early object recognition by designing a state-aware module to predict whether it reaches the stable state. In this way, we can enable accurate recognition with confidence as early as possible. 
To the best of our knowledge, we are the first ones to verify the superiority of event-wise processing in early object recognition.

To summarize, the contributions of this paper are as follows:
\begin{itemize}

   \item We propose a novel graph-based recursive algorithm that enables efficient event-wise processing for event cameras. 

    \item We introduce a novel incremental graph convolution for event-wise processing. It reduces the computational complexity up to 100 times compared to the naive sliding-window-based graph convolution.
    
    \item We propose an event-specific radius search algorithm that reduces query and insertion/deletion costs to make graph construction faster.

    \item Experiments demonstrate that our efficient event-wise algorithm achieves similar performance with batch-wise methods on standard recognition task while enabling early object recognition with confidence.
    
\end{itemize}

\section{Related Work}
    Here we review the existing representation for event-based data in three parts: (1) event-specific representation; (2) event images and voxel grid; (3) point set and graph.

\noindent\textbf{Event-Specific Design.}
    Traditional methods have designed an event-specific representation, namely time surface~\cite{LagorceOGSB17HOTS},
    which is manifested as a 2D map formed with the timestamps of the most recent events. 
    Typically followed by a lightweight model, this representation can be easily updated with each newly arrived event,
    unlocking the low latency advantage of the event camera. Time surface has been applied in different tasks, \eg, stereo event-based SLAM~\cite{stereo-event-slam} and image reconstruction~\cite{image-recon}.
    While many variants~\cite{SironiBBLB18HATS, ManderscheidSBM19SpeedInvari} have been developed, their performance degrades on highly textured scenes~\cite{MuegglerBS17FastCorner} due to the “motion overwriting” problem. Another type of method tailored for event cameras adopts the Spike Neural Network (SNN)~\cite{LiuRXT020spiking_recognition, OrchardBET13spiking_motion, OrchardMEPTB15sipiking_hfirst, AmirTBMMNNAGMKD17spiking_gesture}, which is also bio-inspired designed like event cameras. SNN exploits the sparse and asynchronous nature of events data, but training such networks is difficult due to their non-differentiable character. %

\noindent\textbf{Event Images and Voxel Grid.}
    These methods~\cite{CanniciCRM20matrix-lstm, GehrigLDS19est, Rebecq19cvpre2vid} try to make event data compatible with frame-based technologies. Earlier approaches use simple ways (\eg, counting events or accumulating pixel-wise polarity) to convert the event stream into 2D event frames~\cite{DBLP:conf/ijcnn/CookGJKS11}. Such event frames, which reveal spatial information of scene edges, have been applied to several tasks, \eg, visual odometry~\cite{DBLP:journals/ral/RebecqHGS17}, feature tracking~\cite{DBLP:journals/ijcv/GehrigRGS20}. However, they quantify the timestamp and discard the sparsity property of events data. To improve the temporal resolution, Zhu~\etal~\cite{DBLP:journals/corr/abs-1812-08156Unsupervised, DBLP:journals/corr/abs-1811-08230HighDynamic} suggest discretizing the time dimension into consecutive temporal bins. They accumulate events into a voxel grid through a linearly weighted accumulation similar to bilinear interpolation. Messikommer~\etal~\cite{DBLP:journals/corr/AsynchronousSparse} further exploit spatial and temporal sparsity by adopting sparse convolution~\cite{GrahamEM18Submanifold} and developing a recursive convolution formula. However, their operations are still on sparse volumes. It's challenging for them to process vast event clouds due to the expensive computation cost of 3D convolution.

\begin{figure*}
\begin{center}
\includegraphics[width=1.0\textwidth]{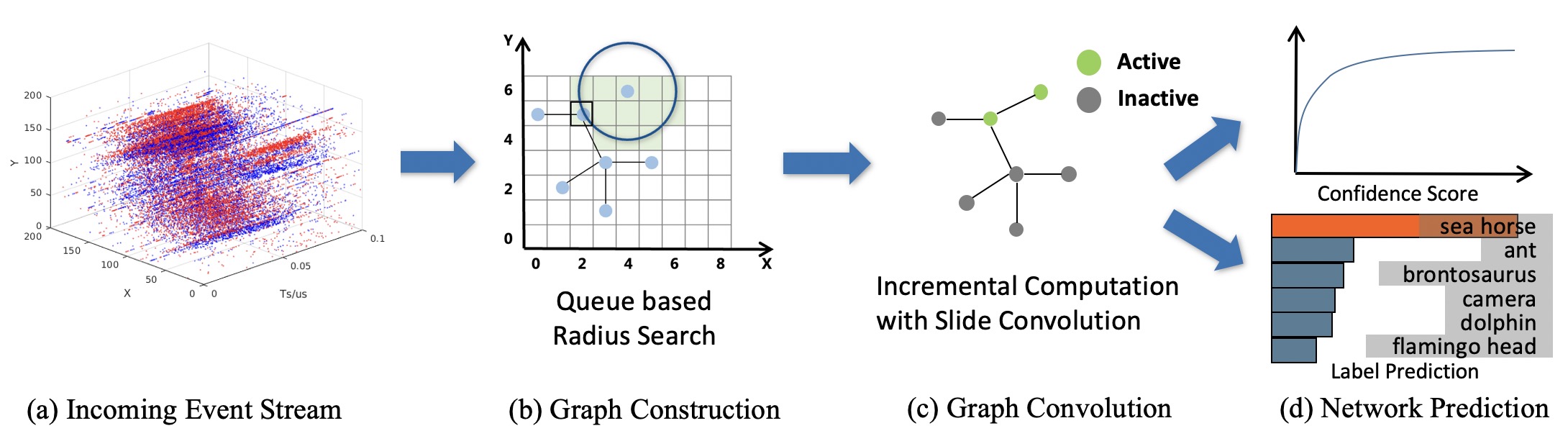}
\end{center}
   \caption{\textbf{Our graph-based Asynchronous Event Processing Framework.} It can efficiently process in an event-wise manner and enable early object recognition, which is mainly thanks to (b) an event-specific radius search algorithm for graph construction, (c) incremental graph convolution for efficient event-wise processing, and (d) bottom branch for object recognition prediction and top branch, \ie, a state-aware module predicting whether it reaches the stable state.
}
\label{fig:pipeline}
\vspace{-1.0em}
\end{figure*}

\noindent\textbf{Point Set and Graph.}
    Ryad~\etal~\cite{benosman2013eventflow} solve optical flow estimation by plane fitting to the event point cloud, an early work that interprets an event sequence as 3D point clouds. Recent works, for example,
    Wang~\etal~\cite{wang2019gesture} further use a PointNet~\cite{qi2017pointnet, qi2017pointnet++}-like framework, which utilizes multi-layer-perceptron to learn features of each point separately, and then outputs object-level responses (\eg, classification labels) through  global max operations. For event-wise processing,
    Sekikawa~\etal~\cite{sekikawa2019eventnet} first develop a recursive architecture, namely EventNet. Specifically, it formulates dependence on causal events to the output recursively using a novel temporal coding and aggregation scheme and pre-computes the node features corresponding to specific spatial coordinates and polarities. However, due to its approximate calculation and the lack of hierarchical architecture, extending EventNet to other high-level tasks is challenging.
    To better exploit the topological structure, \cite{event-graph-cut, bi2019graph} interpret the event cloud in the form of space-time graphs.
    In particular, Bi~\etal~\cite{bi2019graph} show that such compact graph representation requires less computation and memory than conventional CNNs while achieving superior results to the state-of-the-art in various datasets. Mitrokhin~\etal~\cite{mitrokhin2020learning} show that capturing the changes over large time intervals can resolve motion ambiguities. However, such a large time interval will result in a very low response frequency. A recursive formula for graph-based processing needs to be studied, which motivates this paper. %

\section{Preliminaries}

    We first introduce how to build a graph from the event stream. Then we introduce spatial graph convolution, which is the basis of our slide convolution.

\subsection{Event Graph}
    Event cameras respond to changes in the logarithmic brightness signal $L(u_i, t_i) \dot{=} log I(u_i, t_i)$ asynchronously and independently for event pixel~\cite{event_survey}. An event is triggered at pixel $u_i = (x_i, y_i)$ and at time $t_i$ as soon as the brightness increment since the last event at the pixel reaches a threshold $\pm C$ (with $C > 0$):
\begin{equation}
    L(u_k, t_k) - L(u_k, t_k - \Delta t) \geq p_{k}C, 
\end{equation}
   where $p_i\in \{-1,1\}$ is the polarity of the brightness change and $\Delta t$ is the time since the last event at $u_i$. An asynchronous event stream can be expressed as a sequence of events:
\begin{equation}
    \{{\rm event}_{i}\}_B=\{x_i, y_i, t_i, p_i\}_B, 
\end{equation}
    where $B$ is the length of events sequence.
    
    From an event stream, we can construct a graph which is denoted as $G=\{V,E\}$ where $V$ and $E$ represent nodes and edges, respectively. Each event is a node in the event graph, which contains %
    a 3D coordinate $(x_i,y_i,t_i)$ and nodes attribute $(p_i)$. It is also possible to remove or include additional attributes like event-surface normals.
    
    The connectivity of nodes in the graph is usually established by the radius-neighborhood graph strategy. Namely, neighboring nodes $v_i$ and $v_j$ are connected with an edge only if their weighted Euclidean distance $d_{i,j}$ less than radius distance $R$. Before radius search,
    the temporal axis of the event cloud is upscaled by a factor to keep the density of events more uniform across the $x,y,t$ axes. Each edge has its own attribute $e_{ij}$, which is often computed by relative Cartesian coordinates of linked nodes.
    To limit the size of the graph, the connectivity degree for each node is usually constrained to a parameter $D_{max}$.

\subsection{Spatial Graph Convolution}

    Spatial Graph Convolution~\cite{HamiltonYL17spatialconv_sage, abs-1710-10903spatialconv_gat} works by constructing a local neighborhood graph and applying convolution-like operations on the edges connecting neighboring pairs of points.
    Formally, it aggregates a new feature vector for each vertex using its neighborhood information weighted by a trainable kernel function. By using summation as the aggregation operation, it can be defined as:

\begin{equation}
\begin{split}
    (f\otimes g)(i) = \sum_{j\in E(i)} f(j) h_\theta,  \\
    h_\theta = h_\theta(f(i), f(j), e_{ij}), 
\label{eq:spatial-conv}
\end{split}
\end{equation}
    where $\otimes$ is the graph convolution operator, $g$ is kernel function, $f$ is node feature. $E(i)$ is the set of node $i$'s neighbor and $h_\theta$ is a function determining how the features are aggregated by making use of two node features and edge attributes.

\section{Method}
    Inspired by~\cite{sekikawa2019eventnet, DBLP:journals/corr/AsynchronousSparse}, we develop a recursive formula for spatial graph convolution, namely slide convolution. Specifically, slide convolution takes events one by one as input and responds in an event-wise manner while maintaining the structure of the past graph internally (Section~\ref{sec:slide-conv}). To make graph construction faster, we develop a radius search algorithm which better exploits the structure of events cloud against generic 3D continuous metric space (Section~\ref{sec:radius-search}).
    In Section~\ref{sec:state-aware}, we introduce how to apply our graph-based method to early object recognition by combining it with a state-aware module. Fig.~\ref{fig:pipeline} overviews the proposed pipeline.

\begin{figure}
\begin{center}
\includegraphics[width=0.95\linewidth]{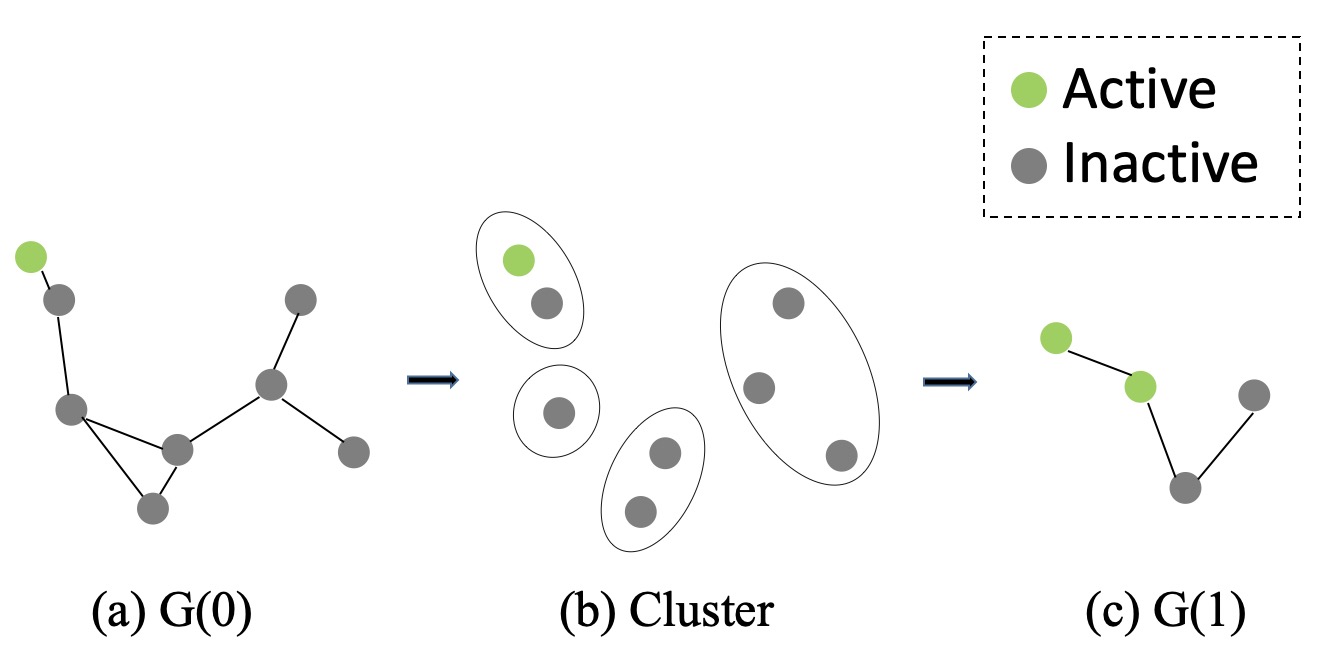}
\end{center}
   \caption{\textbf{An example of the propagation of modified features between different layers.} (a) A newly active node in layer 0 (denoted as G(0)). The active state means that it needs to be updated. (b) Graph pooling which causes a change in the topology. (c) Active nodes in G(1).}
\label{fig:slide-conv}
\end{figure}

\subsection{Slide Convolution} \label{sec:slide-conv}

    To enable spatial graph convolution (denoted as convolution for short in the following) work in an event-wise manner, one straightforward idea is to use a sliding window strategy, i.e., consecutively updating the graph by sliding new events in and sliding events out, then apply convolution on the full graph.
    This way, however, is infeasible because it requires processing the entire window of events again and again at a high event rate.
    A simple way to improve it will be just computing the features around %
    the newly active or inactive nodes (corresponding to the events of sliding in and sliding out). 
    But it only works for the case of single layer, while current modern architecture usually contains multiple layers, which even cause a change in graph topology (Fig.~\ref{fig:slide-conv}).
    Slide convolution solves these problems by deriving a series of propagation rules which helps to propagate the changes from the input layer to deeper layers. The following will focus on how to derive these propagation rules. 
    
    Firstly we rewrite the convolution in a multi-layer architecture as:
    
\begin{equation}
\begin{split}
    f_{n+1}(i) &= \sum_{j\in N(i)} f_{n}(j) h_\theta, for\ i\in A_{n+1}, \\
    \label{eq:sync}
\end{split}
\end{equation}  
    where $f_n$ and $f_{n+1}$ are node features at layer $n$ and $n+1$ respectively (layer 0 is input layer). $A_{n+1}$, namely existing set, represents all existing nodes in the graph at layer $n+1$ (which will change with different sliding windows). $N(i)$ is a map that stores which nodes at layer $n$ contribute to node $i$ at layer $n+1$. Here for convolution, $N(i)$ is one-hop neighbour of node $i$.
    
    Eq.~\eqref{eq:sync} leads of course to redundant computation. We seek to leverage the temporal sparsity of the event stream, \ie, some nodes stay same values at two consecutive timestamp, for efficient computing, which has the following form:
\begin{equation}
\begin{split}
    f_{n+1}^{t+1}(i) &=  f_{n+1}^{t}(i) + \Delta_{n+1}(i), \\
    \Delta_{n+1}(i) &= \sum_{(j,i)\in E_{n+1}}(f_{n}^{t+1}(j) - f_{n}^{t}(j))h_\theta,
    \label{eq:async}
\end{split}
\end{equation}

    The most critical part is $E_{n+1}$, a set of directed edges containing all the edges that point to modified nodes.
    If we know $E_{n+1}$, we can calculate the change of features at time $t+1$ compared to that at time $t$, \ie, $\Delta_{n+1}$ and update the node features.
    
    Notice that for newly active nodes, their states at time $t$ are undefined (similar for newly inactive nodes). In order to distinguish these nodes,
    we divide nodes that need to be updated into three categories: the ones deleted from the graph, the ones newly added to the graph, and the nodes that locate in the receptive fields of these two types of nodes. We use $V^{del}$, $V^{add}$, $V^{up}$ to represent these three kinds of nodes and further split the $E$ into $E^{del}$, $E^{add}$, $E^{up}$ according to which node it points to.
    
    At time $t+1$, for layer 0, $V_{0}^{add}$, $V_{0}^{del}$, $V_{0}^{up}$ is initialized as events sliding into the window, events sliding out of the window and an empty set, respectively. $E_{0}$ is initialized as an empty set. Then we can deduce layer $n+1$ through simple set operations when layer $n$ is given:
    
\begin{equation}
\begin{split}
    V_{n+1}^{add} &= V_{n}^{add}, V_{n+1}^{del} = V_{n}^{del}, \\
    V_{n+1}^{up} &= \{i \mid for\ i \in A_{n+1}^{t}\setminus V_{n+1}^{del},\ if\ \exists j\in N(i)\land j \in V_{n}\}, \\
    V_{n+1} &= V_{n+1}^{add} \cup V_{n+1}^{del} \cup V_{n+1}^{up}, \\
    A_{n+1}^{t+1} &= A_{n+1}^{t} \cup V_{n+1}^{add} \setminus V_{n+1}^{del},
    \label{eq:v-n+1}
\end{split}
\end{equation}

\begin{equation}
\begin{split}
    E_{n+1}^{del} &= \{ (j,i) \mid for\ i\in V_{n+1}^{del}\ then\ \forall j\in N(i)\land j\in A_{n}^{t+1} \}, \\
    E_{n+1}^{add} &= \{ (j,i) \mid for\ i\in V_{n+1}^{add}\ then\ \forall j\in N(i)\land j\in A_{n}^{t+1} \}, \\
    E_{n+1}^{up} &= \{ (j,i) \mid for\ i\in V_{n+1}^{up}\ then\ \forall j\in N(i)\land j\in V_n \}, \\
    E_{n+1} &= E_{n+1}^{add} \cup E_{n+1}^{del} \cup E_{n+1}^{up}, \\
    \label{eq:e-n+1}
\end{split}
\end{equation}    

    In Eq.~\eqref{eq:v-n+1}, $V_{n+1}^{add}$ and $V_{n+1}^{del}$ are directly inherited from last layer since the convolution does not change graph topology. Notice that each time $V_{n+1}$ and $E_{n+1}$ is built from state at previous layer, so we do not need to keep their state at previous moment. $A_{n+1}$, on the contrary, evolved from the state at previous moment, thus we need to distinguish between $A_{n+1}^{t}$ and $A_{n+1}^{t+1} $.
    After reducing $V$ and $E$, we also need to augment those undefined nodes for calculating  $\Delta _{n+1}(i)$.
    Specifically, we expand feature map $f_{n+1}^{t}$ and assign to zeros for $V_{n+1}^{add}$ while for $V_{n+1}^{del}$, we assign $f_{n+1}^{t+1}$ to zeros.
    Now we know how to derive $E_{n+1}$, but we can only deal with convolution. To extend to pooling operations, we need to know the corresponding set of neighborhoods $N(i)$. Take voxel grid pooling as an example. For nodes located in the same voxel (denoted as a set $S(voxel)$), it will be clustered to a  center node. Thus for this center node, its corresponding $N(i)$ is $S(voxel)$.

\begin{figure}[t]
\begin{center}
\includegraphics[width=0.85\linewidth]{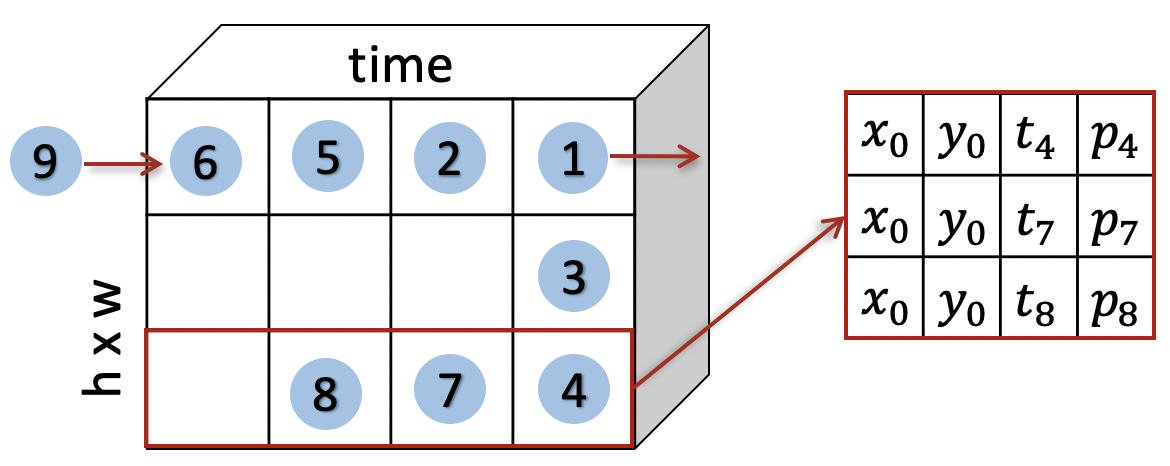}
\end{center}
   \caption{The \textbf{pixel queue} stores events, where each event is a four-tuple. The numbers on the ball represent the order of events (figure adapted from~\cite{TulyakovFKGH19learnseq}).}
\vspace{-1.0em}
\label{fig:pixel-queue}
\end{figure}

    In the supplementary, we prove that using Eq.~\eqref{eq:async} to process events one by one is equivalent to using Eq.~\eqref{eq:sync} to process all events at once. In this way, we can replace spatial convolution with our sliding convolution, leveraging existing graph-based architecture to process events one by one efficiently without sacrificing accuracy. Note that though we use summation for aggregation in the above formulation, it is easy to replace summation by another way, such as max/min aggregation, as long as we know how $E_{n+1}$ evolves.

\subsection{Pixel Queue based Graph Construction} \label{sec:radius-search}
    Recent graph-based methods on event cameras define node connectivity in the graph based on the radius-neighborhood graph strategy, namely radius search. It is usually done by k-d tree, a space-partitioning data structure for organizing points in a k-dimensional space.
    However, we argue that k-d tree does not leverage the structure of the event cloud. 
    What's worse, k-d tree is inefficient when frequent insertions and deletions occur. This is because adding points will make the tree unbalanced, resulting in performance degradation. The same is true for deletion.
    
\begin{figure}[t]
\begin{center}
\includegraphics[width=0.85\linewidth]{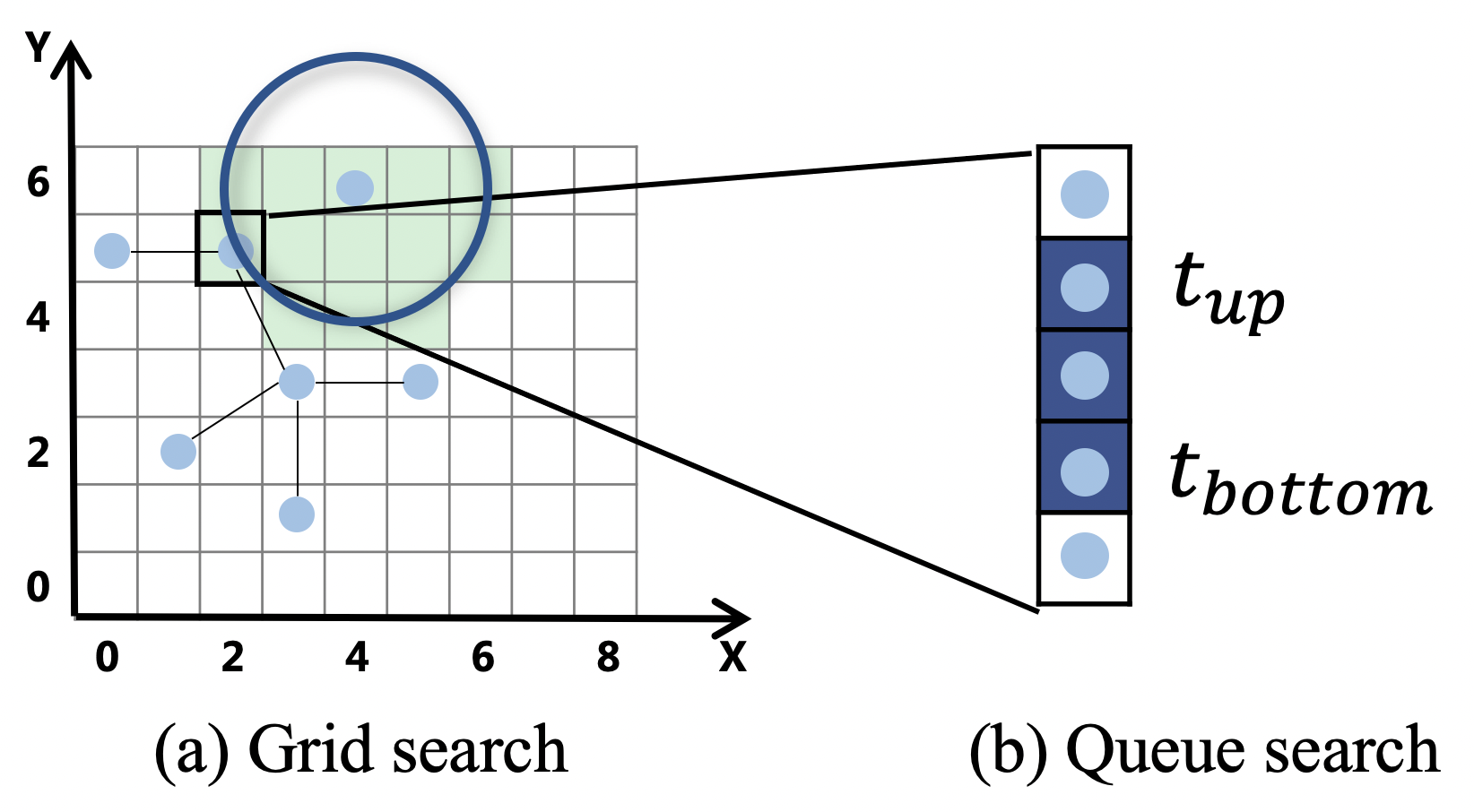}
\end{center}
   \caption{\textbf{A two-stage radius search based on pixel queue}. To search events given the radius, we first determine candidate pixels in the image grid, as shown in (a). Candidate pixels are represented in lime.
   Secondly, for each pixel queue, we collect events between lower bounds $t_{bottom}$ and upper bounds $t_{up}$, as shown in (b).}
\label{fig:rad-search}
\vspace{-2.0em}
\end{figure}

    Instead of k-d tree, we employ pixel queue (See Fig.~\ref{fig:pixel-queue}) to store events, which is an event-specific data structure that has been used in a number of works~\cite{TulyakovFKGH19learnseq, stereo-event-slam, event-graph-cut}. Pixel queue stores the most recent events at each location sorted by the time of their arrival. Based on pixel queue, we propose a two-stage radius search algorithm (see Fig.~\ref{fig:rad-search}).
    The first step is to search in the image grid and filter out the candidate pixel queues that contain the events we want. It can be done with the help of distance field, which describes how far the other pixels away from an anchor pixel. The distance field is similar to the partitioning technology used in k-d tree but only needs to be calculated once.
    For a query event $(x_0, y_0, t_0)$ and the radius $R$, we can determine candidate pixels and corresponding queues whose spatial distance is smaller than the radius by looking up the pre-computed distance field.
    
    In the second step, we traverse these candidate pixel queues. For a candidate pixel queue with spatial offset $(\delta x, \delta y)$ from the query event, the target events contained in it must have lower bound $t_0-\sqrt{R^2-(\delta x^2 + \delta y^2)}$  (denoted as $t_{bottom}$) and upper bound $t_0+\sqrt{R^2-(\delta x^2 + \delta y^2)})$ (denoted as $t_{up}$). We get these events by finding the index of $t_{up}$ and $t_{bottom}$ using binary search. Collecting events in all candidate pixel queues gives a final query result. 
    We evaluate the performance of our method compared to the k-d tree based method in Section~\ref{sec:tra-obj-reg} and analyze the computation complexity in the supplementary.

\subsection{State-aware Module}
\label{sec:state-aware}

    With the increase of input events and information, the prediction result will be stable from a certain moment. By this time, it is meaningless to process more events, and the system should give an early recognition result.
    We fulfill this goal
    with the help of our state-aware module, \ie, top branch in Fig.~\ref{fig:pipeline}-(d). Specifically, we use a multi-layer perceptron, \ie MLP, to represent a state-aware function which maps the graph feature map to a binary prediction. The prediction result means whether it achieves the stable status.
    Then during inference, we interpret the value after activation as the confidence score.
    
    Given the object recognition branch pre-trained, we can generate ground truth labels for training state-aware module. One possible way is to analyze prediction as a function of event index. When the prediction does not change with event index increasing, we consider it stable and define the ground truth as one, otherwise, we define it as zero. For simplification, we adopt an approximate approach, \ie, considering a prediction stable if it is equal to that at the last event index. The ground truth will be compared with network prediction by Binary Cross Entropy with Logits Loss.
    As for the training data, we randomly crop sequences to variable lengths (from 5ms to 50ms) and hope it learn to predict corresponding confidence in different states. 
    It is noteworthy that we do not crop sequences when we train the object recognition branch.

\section{Experiments}

\begin{table*}[htbp]
    \centering
    \begin{tabular}{llllllllll}
    \hline
     &  & \multicolumn{2}{c}{N-Caltech101} & \multicolumn{2}{c}{CIFAR10-DVS} & \multicolumn{2}{c}{MNIST-DVS} & \multicolumn{2}{c}{N-Cars} \\ \cline{3-10}
    Methods & Representation & Acc $\uparrow$ & Mps/ev $\downarrow$ & Acc $\uparrow$ & Mps/ev $\downarrow$ & Acc $\uparrow$ & Mps/ev $\downarrow$ & Acc $\uparrow$ & Mps/ev $\downarrow$ \\ \hline
    H-First~\cite{OrchardMEPTB15sipiking_hfirst} & Spike & 0.054 & - & 0.077 & - & 0.595 & - & 0.561 & \\
    Gabor-SNN~\cite{gabor_filter, SironiBBLB18HATS} & Spike & 0.196 & - & 0.245 & - &  0.824 & - & 0.789 & -   \\
    HOTS~\cite{LagorceOGSB17HOTS} & TimeSurface & 0.210 & 54.0 & 0.271 & \cellcolor{third}26 & 0.803 & 26 & 0.624 & 14.0 \\
    HATS~\cite{SironiBBLB18HATS} & TimeSurface & 0.642 & \cellcolor{first}4.3 & 0.524 & \cellcolor{first}0.18 & 0.984 & \cellcolor{first}0.18 & 0.902 & \cellcolor{first}0.03 \\
    DART~\cite{RameshYOTZX20dart} & TimeSurface & 0.664 & - & \cellcolor{third}0.658 & - & 0.985 & - & - & - \\
    YOLE~\cite{CanniciCRM19yole} & VoxelGrid & \cellcolor{third}0.702 & 3659 & - & - & 0.961 & - & \cellcolor{third}0.927 & 328.16 \\
    Asynet~\cite{DBLP:journals/corr/AsynchronousSparse} & VoxelGrid & \cellcolor{second}0.745 & 202 & \cellcolor{second}0.663 & 103 & \cellcolor{first}0.994 & 112 & \cellcolor{first}0.944 & 21.5 \\
    \hline
    NVS-B (Ours) & Graph & 0.670 & 221 & 0.602 & 601 & \cellcolor{third}0.986 & 154 & 0.915 & 57.9 \\
    NVS-S (Ours) & Graph & 0.670 & \cellcolor{second}7.8 & 0.602 & \cellcolor{second}22.8 & \cellcolor{third}0.986 & \cellcolor{second}10.1 & 0.915 & \cellcolor{second}5.2 \\
    EvS-B (Ours) & Graph & \cellcolor{first}0.761 & 1152 & \cellcolor{first}0.680 & 3020 & \cellcolor{second}0.991 & 548 & \cellcolor{second}0.931 & 251 \\
    EvS-S (Ours) & Graph & \cellcolor{first}0.761 & \cellcolor{third}11.5 & \cellcolor{first}0.680 & 33.2 & \cellcolor{second}0.991 & \cellcolor{third}15.2 & \cellcolor{second}0.931 & \cellcolor{third}6.1 \\
    \hline
    \end{tabular}
    \newline
    \caption{\textbf{Comparison with different representations for object recognition.} We color code each row as \colorbox{first}{best}, \colorbox{second}{second best
    }and \colorbox{third}{third best}. *-B means baseline and *-S means SlideGCN. Our graph-based baseline (EvS-B) achieves the state-of-the-art-performance (i.e. 0.761 on N-Caltech101 and 0.680 on CIFAR10-DVS). Replacing by our slide convolution, the computational complexity reduces  up to two orders of magnitude (1152 vs. 11.5 and 3020 vs. 33.2). As a result, our method (EvS-S) strikes a balance between event-specific low-latency and high-performance high-latency methods.}
    \label{tab:comp-sota}
\end{table*}

\begin{figure*}
\begin{center}
\includegraphics[width=0.95\linewidth]{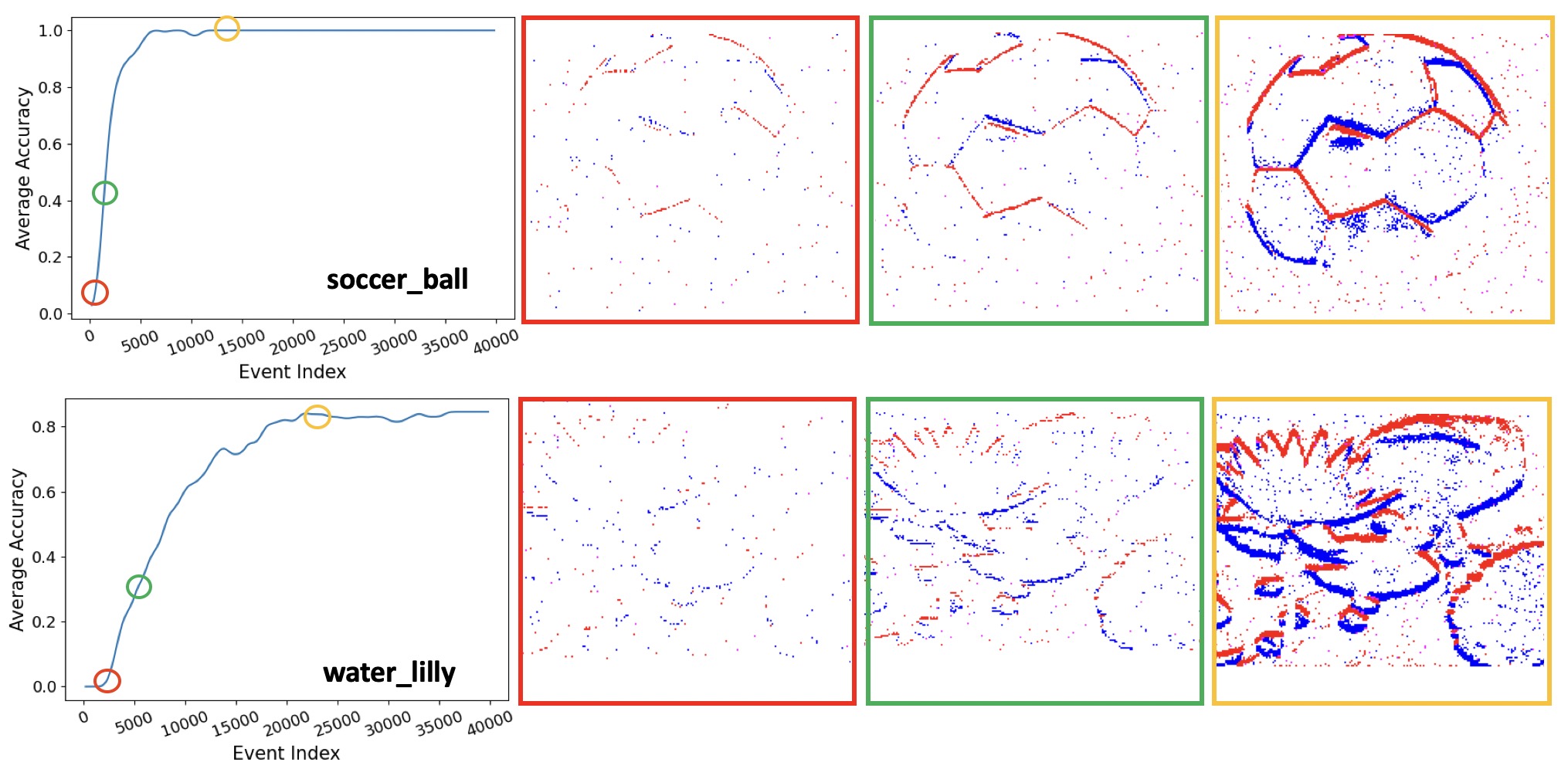}
\end{center}
\vspace{-1.0em}
   \caption{\textbf{Qualitative results of stream-based object recognition.} The accuracy improves and stabilizes as events accumulate. We selected points at three moments in this process, which are circled by red, green, and orange.}
\label{fig:qulitative}
\vspace{-1.5em}
\end{figure*}

\subsection{Object Recognition}\label{sec:tra-obj-reg}

\noindent \textbf{(1) Dataset and Evaluation Metrics.}
    Event-based object recognition is to predict the category of the object through the input event stream. We verify our method's superiority on four datasets, including N-Caltech101~\cite{ncaltech101}, N-Cars~\cite{SironiBBLB18HATS},  CIFAR10-DVS~\cite{cifar10}, and MNIST-DVS~\cite{serrano2015mnist-dvs}.
    
    N-Caltech101, CIFAR10-DVS, and MNIST-DVS are converted from standard frame-based datasets. This is done by displaying a moving image on a monitor and recording with a fixed event camera or fixing the monitor instead of the camera.
    The same as the original Caltech101, N-Caltech101 contains 8246 samples and 101 categories. 
    CIFAR10-DVS, on the contrary, randomly select one-sixth of the original frame-based image dataset, thus containing 6,000 samples per class and 60,000 samples in total. MNIST-DVS consists of 10,000 symbols sampled from the standard MNIST 70,000-picture database, with each of the 10,000 symbols displayed at three different scales, thus containing 30,000 samples in total.
    Different from the former, N-Cars are created by directly recording objects in real-world environments with an event camera. N-Cars comprises two class labels, namely 12,336 car samples and 11,693 non-car samples (background). We sample some sequences from these datasets for visualization in the supplementary material.
    
    To evaluate task performance and the potential ability of event-by-event processing, we consider two metrics: prediction accuracy and floating point operations per second (FLOPs). While the first indicates the quality of the prediction, the second shows the computational complexity required for each event update.
    
\noindent \textbf{(2) Implementation Details.}
    We implement two graph convolution networks to evaluate our SlideGCN. %
    Adapting from~\cite{bi2019graph}, our first architecture, namely NVS, consists of two parts: a backbone and a prediction head. The backbone is comprised of 4 ``GraphConv-ELU-Bn" layers, where “GraphConv” will be replaced by our slide convolution during inference. The prediction head comprises one fully connected layer to map the features to classes. In order to reduce overfitting, a dropout layer with a probability of 0.3 is added after the first fully connected layer. Compared with the original architecture, we replace the cluster-level pooling layer with a readout function, which summarizes the graph-level representation by taking both the max/mean of hidden representations of sub-graphs~\cite{gnn-survey}. 
    The second architecture, which is inspired by~\cite{mitrokhin2020learning}, is named EvS .
    Following the idea from~\cite{mitrokhin2020learning}, we used two constraints when computing edges for EvS. One of them is to keep only the points lying in the upper (along the temporal axis) hemisphere of a point, and the other is to filter the edges so that they are parallel to the event surface.
    This preserves most of an event’s temporal motion information while obtaining sparser edges. We also add the normal of the event surface to the input.
    We use these two networks as our baseline and then replace their convolution layers with our slide convolution during evaluation. %
    Please refer to the supplementary material for more details about the parameters like time interval and network depths.

\noindent \textbf{(3) Comparison to the State-of-the-Art.}
    Table~\ref{tab:comp-sota} compares our results with other state-of-the-art methods. All these methods are able to process event stream event-by-event.
    Thanks to the effective expression of the graph structure, the EvS(baseline) we implemented has reached state-of-the-art on the challenging datasets N-Caltech101 and CIFAR10-DVS. The graph convolution layer is further replaced with our slide convolution, namely EvS(SlideGCN), which reduces the computational complexity up to 100 times without sacrificing the original performance. Another method, \ie, NVS does not perform on par with state-of-the-art, but it is lightweight and requires less computation.
    
    Our method strikes a balance between event-specific low-latency and high-performance high-latency methods. %
    On the one hand, it has less calculation (11.5 vs. 202 on N-Caltech101 and 33.2 vs. 103 on CIFAR10-DVS) than other second-best methods. On the other hand, it achieves 15.6\% higher accuracy on N-Caltech101 and 29.7\% higher accuracy on CIFAR10-DVS, than hand-crafted and event-specific methods, \ie, HATS~\cite{SironiBBLB18HATS}, which have a low-level computation complexity.
    
    Besides the computation complexity, we timed our experiments conducted on N-Caltech101 by measuring the processing time for each event update on an i7-9700K CPU (using a single core). Our method requires 16.9 ms, while the baseline needs 130.4 ms. Therefore, our method is roughly 8 times faster by reusing previous calculations.
    We expect that our method will significantly reduce the running time on the GPU or specific hardware as its lower number of FLOPs.
    Please refer to our supplementary material for more comparisons with the methods that cannot efficiently process event data.

\begin{figure}
\begin{center}
\includegraphics[width=0.95        \linewidth]{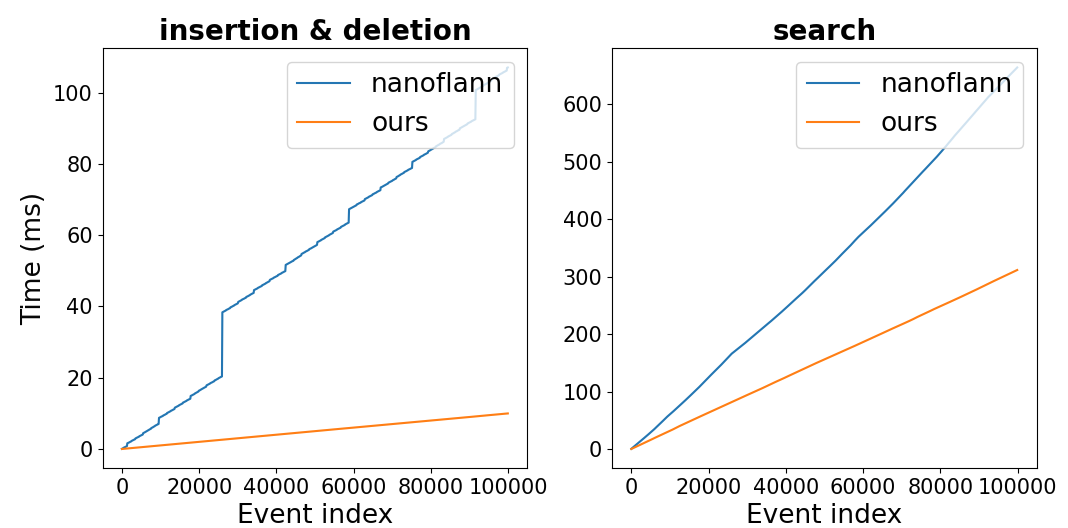}
\end{center}

    \vspace{-0.5em}
   \caption{\textbf{Comparison of our radius search method and k-d tree based method} (with nanoflann implementation).} 
\label{fig:queue-performance}
\vspace{-1.0em}
\end{figure}

\noindent \textbf{(4) Efficiency of Pixel Queue based Radius Search.}
    We evaluate the efficiency of our pixel queue based radius search with nanoflann~\cite{blanco2014nanoflann} , which is a popular k-d tree implementation supporting dynamic update. Specifically, we use a window of 100,000 events, and each time slide by 100 events,
    followed by a radius search on the newly sliding-in events. We repeat sliding 1,000 times and show the cumulative cost of insertion, deletion, and searching in Fig.~\ref{fig:queue-performance}.
    
    The insertion and deletion cost of nanoflann is an order of magnitude higher than our method on average, and it rises rapidly sometimes. That is because nanoflann uses lazy deletion, which does not rebuild the index immediately after removing elements. The cost will have a significant rise when it rebuilds the index.
    As for searching, our method reduces the time cost by half by leveraging the spatial locality of the event cloud.

\subsection{Stream based Object Recognition}\label{sec:stream-obj-reg}

\begin{table}[htbp]
    \begin{tabular}{c|cccc}
    \cline{1-4}
    Method & Batch size (ms) & Accuracy & Index \\ \cline{1-4}
    \multirow{5}{*}{Batch-wise} & 10 & 0.362 & 3829 \\
     & 20 & 0.490 & 7611 \\
     & 30 & 0.615 & 11041 \\
     & 40 & 0.718 & 14771 \\
     & 50 & 0.761 & 19154 \\ \cline{1-4}
    SlideGCN & $12^{*}$ & 0.669 & 3118 \\ \cline{1-4}
    \end{tabular}%
    \newline
    \caption{\textbf{Comparison of SlideGCN and batch-wise method on stream based object recognition.} $12^{*}$ means that time consuming is 12 ms, not batch size.
    }
    \label{tab:stream-comp}
    \vspace{-1.0em}
\end{table}

\begin{table}[htbp]
    \small
    \centering
    \begin{tabular}{c|ccccc}
    \cline{1-5}
    Methods            & Size & Cum MFLOPS & Avg MFLOPS & Index & \\ \cline{1-5}
    \multirow{3}{*}{SlideGCN}           & 1    & 17711 & 5.68 & 3118 & \\
                       & 10   & 8936 & 28.58 & 3127 &\\
                       & 100  & 3041 & 95.64 & 3170 &\\ \cline{1-5}
    Batch-wise         & -    & 1152 & - & 19154 & \\ \cline{1-5}
    \end{tabular}
    \newline
    \caption{\textbf{Cumulative MFLOPS with different mini-batch sizes.}}
    \label{tab:trade-off}
    \vspace{0.0em}
\end{table}

We designed the stream based object recognition task to verify the effectiveness of our event-wise processing. In this task, we evaluate the prediction accuracy when the algorithm claims it gives a reliable response. Specifically, we consider it a reliable result for our event-wise method when our state-aware module provides a high confidence score. While for the batch-wise method, we consider that the confidence of each processing is equal to one. In this way, we measure the accuracy and the latency at the same time.

\noindent \textbf{(1) Comparison to the batch-wise way.}
    We use a window configuration of 50 milliseconds to train the network. Using the same network, we test in the batch-wise way and event-wise way (by replacing with our slide convolution) separately.
    As shown in Table~\ref{tab:stream-comp}, for the batch-wise way, decreasing batch size reduces latency but also causes drops in accuracy. On the contrary, our SlideGCN performs close accuracy (0.669 vs. 0.761) with the best configured batch-wise manner but response much earlier (3118 vs. 19154 in terms of event index and  12ms vs. 50ms in terms of time). 
    Here we analyze how our method works.
    In %
    Fig.~\ref{fig:qulitative} we show how the accuracy increases with cumulative events for two kinds of objects. The curves vary because the texture richness of the objects is different. Different kinds of motion also cause discrepancies. As a result, it is not trivial to tune a perfect batch size for batch-wise methods. While choosing a big batch size ensures high accuracy (a big batch size means that it receives enough information for most of the objects), it requires too many events as input. Choosing a small batch size, on the contrary, can not guarantee to receive enough information for many objects, thus resulting in low accuracy. Unlike the batch-wise way, our method is not limited to fixed batch size and works in an event-wise manner. Combined with our state-aware module, it can process event-by-event, predict a confidence score simultaneously. As soon as it detects a stable state with high confidence, we can stop processing the following events and make an early recognition.

\noindent \textbf{(2) Trade-off between latency and computational effort.}    
    In practice use, there is a trade-off between the latency and the computation load. Event-wise processing minimizes the latency of the data, but it also makes the computation load very large. We declare that our method is not limited to event-wise processing but can be extended to mini-batch and batch-wise. In Table~\ref{tab:trade-off} we compare
    cumulative MFLOPS (denoted as Cum MFLOPS) with different mini-batch sizes. 
    The table shows that although the event-wise method owns the lowest latency, it causes higher cumulative FLOPs. Increasing the mini-batch size will reduce cumulative FLOPs but in the cost of bringing more latency (an extreme case is that the entire window is used as a batch). In practical use, we can set a mini-batch size of 100 to achieve a balance between computation load and latency.

\section{Conclusion}
In this paper, we introduce a novel graph-based recursive algorithm for event cameras, which is able to keep a high performance of graph convolution networks as well as the ability  of event-by-event processing. To achieve this, we propose a novel incremental convolution method that significantly reduces computational complexity compared to the naive sliding window strategy. To make graph construction faster, we also exploit the structure of the events cloud and develop an event-specific radius search algorithm based on  pixel  queue. The experiments demonstrate that our efficient event-wise algorithm achieves similar performance with batch-wise methods on standard recognition task while enabling early object recognition with confidence.

\clearpage

{\small
\bibliographystyle{ieee_fullname}
\bibliography{main}
}

\end{document}